\documentclass[lettersize,journal]{IEEEtran}
\usepackage{amsmath,amsfonts}
\usepackage{algorithmic}
\usepackage{algorithm}
\usepackage{array}
\usepackage{textcomp}
\usepackage{multirow}
\usepackage{stfloats}
\usepackage{url}
\usepackage{verbatim}
\usepackage{graphicx}
\usepackage{adjustbox}
\usepackage{mathtools}
\usepackage{makecell}
\usepackage{amssymb}
\usepackage{cite}
\usepackage{color}
\usepackage{subfigure}

\usepackage{hyperref}
\hypersetup{
    colorlinks=true,
    linkcolor=blue,
    filecolor=magenta,      
    urlcolor=blue,
}

\begin{document}

\title{Incorporating Point Uncertainty in Radar SLAM}

\author{Yang Xu$^\dag$, Qiucan Huang$^\dag$, Shaojie Shen, and Huan Yin
\thanks{$^\dag$ Yang Xu and Qiucan Huang contributed equally to this study. }
\thanks{All authors are with the Cheng Kar-Shun Robotics Institute, Hong Kong University of Science and Technology, Hong Kong SAR.}
\thanks{This work was supported in part by the HKUST-DJI Joint Innovation Laboratory, and the Hong Kong Center for Construction Robotics (InnoHK center supported by Hong Kong ITC).  \textit{(Corresponding Author: Huan Yin)}}
}

\markboth{Accepted by IEEE Robotics and Automation Letters}%
{Shell \MakeLowercase{\textit{et al.}}: A Sample Article Using IEEEtran.cls for IEEE Journals}


\maketitle

\begin{abstract}
Radar SLAM is robust in challenging conditions, such as fog, dust, and smoke, but suffers from the sparsity and noisiness of radar sensing, including speckle noise and multipath effects. This study provides a performance-enhanced radar SLAM system by incorporating point uncertainty. The basic system is a radar-inertial odometry system that leverages velocity-aided radar points and high-frequency inertial measurements. We first propose to model the uncertainty of radar points in polar coordinates by considering the nature of radar sensing. Then, the proposed uncertainty model is integrated into the data association module and incorporated for back-end state estimation. Real-world experiments on both public and self-collected datasets validate the effectiveness of the proposed models and approaches. The findings highlight the potential of incorporating point uncertainty to improve the radar SLAM system. We make the code and collected dataset publicly available at \href{https://github.com/HKUST-Aerial-Robotics/RIO}{https://github.com/HKUST-Aerial-Robotics/RIO}.
\end{abstract}


\section{Introduction} \label{sec:Introduction}
\IEEEPARstart{K}{nowing} one's own pose is a fundamental problem for robotics and the navigation system. Recent state estimation techniques, such as simultaneous localization and mapping (SLAM), are widely used for pose estimation for navigation systems. Advancements in sensing technology have promoted the development and real-world deployment of visual and laser-based SLAM, either independently or through sensor fusion approaches. These sensing modalities might fail well in adverse conditions, such as smoke scenarios caused by indoor fire scenes or outdoor snowy environments, thus blocking the application of mobile robots in these demanding situations.

Radio detection and ranging (Radar) is a robust all-weather sensing modality capable of operating effectively under challenging conditions~\cite{hong2022radarslam,yin2021rall}. On the other hand, radar measurements are typically sparse and noisy points, which pose significant challenges for front-end data processing and back-end pose estimation. To tackle this, researchers have proposed approaches to enhance the accuracy and robustness of radar-based SLAM, such as the integration of Doppler velocity information~\cite{kramer2020radar,Huang2024Less} and scan-to-map matching techniques~\cite{zhuang20234d,kubelka2023we}. We also observe that existing radar SLAM systems often fuse radar points with inertial measurements for pose tracking, thereby advancing the utility of radar SLAM in practical robotic applications.

Generally, SLAM systems involve not only pose but also uncertainty. The uncertainty spans from front-end measurements to back-end states, describing the ``confidence'' of measurements and states. However, most existing radar SLAM systems mainly focus on motion (state) estimation, i.e., offering sequential poses with the given input. We consider that incorporating the uncertainty of radar measurements could be a key to performance enhancement. Thus two questions are naturally raised: \textit{How do we model the uncertainty of radar measurements?} and \textit{How do we enhance the performance of radar SLAM with the uncertainty modeling?}

This study tries to answer the two questions above. Specifically, we adopt our previously designed radar-inertial odometry (RIO)~\cite{Huang2024Less} as a reduced yet basic radar SLAM. The system takes advantage of velocity-aided radar points and high-frequency inertial measurements. Then we propose to model the uncertainty of radar points in polar coordinates by considering the nature of radar sensing. The modeled uncertainty is leveraged in data association in a probabilistic manner. Additionally, weighted least squares are built based on the proposed uncertainty modeling, thus improving the radar SLAM performance for motion estimation. We conduct real-world experiments on the public Coloradar dataset~\cite{kramer2021coloradar} and our mobile platform with two different radar sensors. Overall, our key contributions in this study are summarized as follows,
\begin{itemize}
    \item Modeling uncertainties for 3D radar measurement model in polar coordinates.
    \item Incorporating the modeled uncertainty in data association and weighted least squares to enhance performance.
    \item Conducting ablation studies and comparisons on real-world datasets for validation.
    \item Open-sourcing the code and collected datasets to benefit the community.
\end{itemize}



\section{Related Work} \label{sec:RelatedWorks}

\subsection{Uncertainty modeling in SLAM}

Uncertainty is a fundamental term in SLAM. The modeled uncertainty could be used for various SLAM-related topics, such as active SLAM~\cite{rodriguez2018importance} and motion detection in dynamic environments~\cite{yin2022dynam}.

Modeling uncertainty is the condition before we leverage it into SLAM systems. Typical SLAM systems are based on the Gaussian assumption~\cite{thrun2002probabilistic,barfoot2024state}, hence the uncertainty is associated with the covariance matrix. Visual sensing and perception are generally data-rich, and neural networks could be trained to estimate the covariance matrices. Shan \textit{et al.}~\cite{shan2020orcvio} propagated such uncertainty into the visual-inertial odometry. In the object SLAM system by Merrill \textit{et al.}~\cite{merrill2022symmetry}, the uncertainty was estimated by trained neural networks and utilized for outlier rejection.

Increased interest is shown in the covariance estimation in recent LiDAR-inertial odometry (LIO) systems. Unlike visual perception, data-driven range sensing is not so mature. Thus, related studies mainly focus on model-based covariance estimation. Yuan \textit{et al.}~\cite{yuan2021pixel} designed a LiDAR-based voxel mapping scheme, in which a plane-based uncertainty model is proposed in discrete voxels. The experimental results demonstrate that using the uncertainty model could improve the mapping precision. In the work proposed by Jiang \textit{et al.}~\cite{jiang2022lidar}, the uncertainty was quantified on point-based tree structures without space discretization, making it applicable to uneven laser point clouds. The tree structure is built on the nearest neighbor search LiDAR points. This study is inspired by these existing research works. Though LiDAR and radar are both range sensings, measurements from system-on-chip radar are generally sparser than those from LiDAR sensors. Thus, directly applying existing uncertainty modeling of dense LiDAR points is inappropriate.


\subsection{Radar SLAM for Motion Estimation}

Radar-based SLAM has a long history, dating back to the late 20th century~\cite{clark1999simultaneous}. While radar is robust in all types of weather, it is often noisy and sparse, bringing difficulties to the state estimation for robotics. To address this, recent studies have suggested combining radar data with high-frequency inertial data to enhancerobustness and accuracy. Doer and Trommer~\cite{doer2020ekf} used an Extended Kalman Filter (EKF) for their RIO system. Kramer \textit{et al.}~\cite{kramer2020radar} proposed to fuse the sensor data via sliding-window optimization, and the Doppler velocity of radar was also considered. Michalczyk \textit{et al.}~\cite{michalczyk2023multi} designed a multi-state and tightly-coupled system, in which radar points served as landmarks for data matching. In 4D iRIOM~\cite{zhuang20234d}, Doppler velocity was used to filter out radar noise, and a technique common in LiDAR SLAM, known as distribution-to-multi-distribution matching, was developed for estimating motion. Differently, Lu \textit{et al.}~\cite{lu2020milliego} introduced a data-driven-based scheme for sensor fusion. Most of the work mentioned above mainly focuses on motion estimation by regarding radar data as 3D points without considering the uncertainty of measurements. The previous work \cite{kellner2014instantaneous} by Kellner \textit{et al.} considered uncertainty in 2D angles towards the location for Doppler velocity residuals. In this study, we build a complete sensor fusion framework and incorporate the point uncertainty in it.


\section{Methodolody} \label{sec:Methodology}

\begin{figure}[t]
    \centering
    \includegraphics[width=0.95\columnwidth]{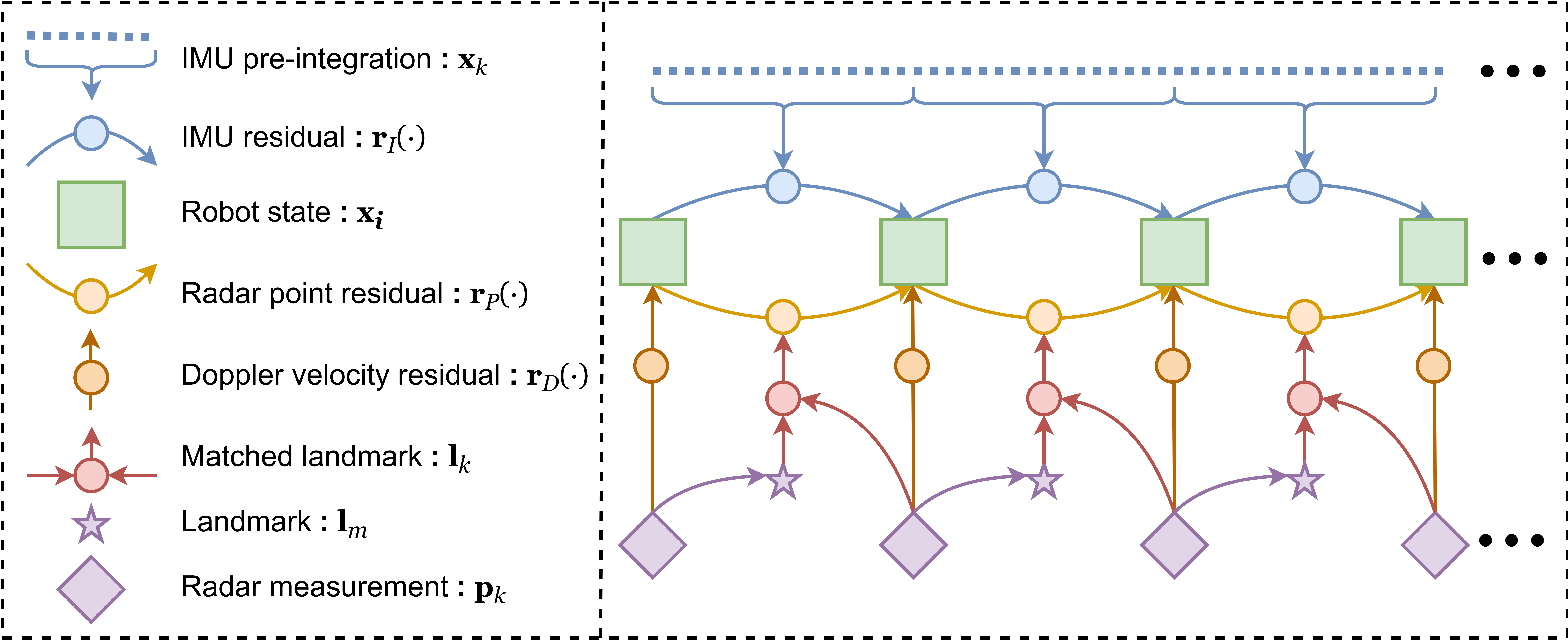}
    \caption{Factor graph representation of our radar-inertial odometry. Three residuals as factors are constructed to estimate robot poses and map landmarks.} 
    \label{fig:system}
\end{figure}

\subsection{System Overview} \label{sec:Overview}

The complete state of our estimation problem is defined as follows:
\begin{equation}
\begin{aligned}
    &\mathbf{X} = \left[ \mathbf{x}_{0}^{T}, \mathbf{x}_{1}^{T}, \cdots ,\mathbf{x}_{k}^{T}, \mathbf{l}_{0}^{T}, \mathbf{l}_{1}^{T}, \cdots ,\mathbf{l}_{n}^{T}  \right]^{T} \\
    &\mathbf{x}_{i} = \left[ \mathbf{p}_{i}^{T}, \mathbf{q}_{i}^{T}, \mathbf{v}_{i}^{T}, {\mathbf{b}_{a}}_{i}^{T}, {\mathbf{b}_{g}}_{i}^{T}  \right]^{T}, \quad i \in \left[0, k \right] \\
    &\mathbf{l}_{j} \in \mathbb{R}^{3}, \quad j \in \left[ 0, n \right] 
\end{aligned}
\end{equation}
in which $\mathbf{X}$ contains robot states $\mathbf{x}_{i}$ and landmarks $\mathbf{l}_{j}$. Specifically,  $\mathbf{p}_{i}$, $\mathbf{q}_{i}$ and $\mathbf{v}_{i}$ represent the position, rotation, and velocity of the $i$-th frame, respectively. The terms ${\mathbf{b}_{a}}_{i}$, ${\mathbf{b}_{g}}_{i}$ are accelerometer bias and gyroscope bias in the $i$-th frame from inertial measurement unit (IMU). The term $\mathbf{l}_{j}$ is the $j$-th map landmark.

The entire system is formulated as a sliding window-based optimization framework that aims to minimize a set of residuals assigned with a Mahalanobis norm, as follows:
\begin{equation} \label{eq:OptProblem}
\begin{aligned} 
    \min_{\mathbf{X}}\bigg\{ \sum_{k \in I} \lVert\mathbf{r}_{I}(\mathbf{x}_{k},\mathbf{x}_{k+1}) \rVert_{\mathbf{\Sigma}_{k}^{k+1}} + & \sum_{k \in D} \lVert \mathbf{r}_{D}(\mathbf{x}_{i},p_{k}) \rVert_{\mathbf{\Sigma}_{k}} + \\
    &\sum_{k \in P} \lVert \mathbf{r}_{P}(\mathbf{x}_{i},\mathbf{l}_{k},p_{k}) \rVert_{\mathbf{\Sigma}_{k}} \bigg\}
\end{aligned}
\end{equation}
where $p$ represents a radar point. The terms $\mathbf{r}_{I}(\cdot)$, $\mathbf{r}_{D}(\cdot)$ and $\mathbf{r}_{P}(\cdot)$ are three different residuals; $\mathbf{I}$, $\mathbf{D}$, and $\mathbf{P}$ encodes the set of IMU, Doppler velocity, and point measurements, respectively; $\lVert \cdot \rVert_{\Sigma}$ represents the Mahalanobis norm. Figure~\ref{fig:system} presents the factor graph-based visualization of the system. The formulation is consistent with the system design in our previous work~\cite{Huang2024Less}.

Specifically, in Equation~\eqref{eq:OptProblem}, the first residual term, $\mathbf{r}_{I}(\cdot)$, is the IMU propagation for estimating continuous radar states. It provides an initial guess for optimizing the entire system. An IMU-only dead-reckoning system could not guarantee the accuracy and reliability of robotic motion estimation. Thus we also propose to leverage radar sensing into the optimization, resulting in the other two residual terms $\mathbf{r}_{D}(\cdot)$ and $\mathbf{r}_{P}(\cdot)$ that will be introduced in the subsequent Section~\ref{sec:DopplerConstraint}, Section~\ref{sec:ProbMatching} and Section~\ref{sec:PointConstraint}, respectively.


\subsection{Radar Measurement Model} \label{sec:NoiseModel}

Before delving into the residual functions, it is imperative to address the modeling of radar measurements and the associated noise, i.e., uncertainty modeling. This modeling facilitates an understanding of the inherent uncertainty in radar measurements, thereby enhancing the accuracy of motion estimation through the adjustment of residual weights.


Radar sensors emit observations as azimuth and elevation angles, along with the detected range. As pointed out by the previous study~\cite{retan2022radar}, employing polar coordinate measurement models offers a more accurate formulation of noise covariance compared to the conventional Cartesian model. Given a radar point $\prescript{R}{}{p_k}$ in the radar frame $R$, we decompose $\prescript{R}{}{p_k}$ to two distinct components: the range measurement, represented by $r_k \in \mathbb{R}$; and the azimuth and elevation measurements on the bearing direction, denoted as $\mathbf{\Omega}_k \in \mathbb{S}^2$.

Regarding the range noise, it can be modeled as a normal distribution with a mean of zero and a standard deviation of $\sigma_r$, indicated as: $\delta_{r_k} \sim \mathcal{N}(0, \sigma_r^2)$. Specifically, $\sigma_r$ is the standard deviation of the range measurement noise, encapsulating the variability of the distance measurements around their true value within the defined confidence interval. The ground-truth range can be represented as:
\begin{equation} \label{r_noise}
r^{gt}_k = r_k + \delta_{r_k}
\end{equation}
\par For the angular measurements, we model the azimuth and elevation noise as independent Gaussian distributions centered around zero with standard deviations. Thus the angular measurement noise of $\mathbf{\Omega}$ is represented as: $\delta_{\mathbf{\Omega}_k} \sim \mathcal{N}\left(\mathbf{0}_{2\times1}, \mathbf{\Sigma}_{\mathbf{\Omega}}\right)$. The term $\mathbf{\Sigma}_{\mathbf{\Omega}}$ is the covariance matrix for the angular measurements:
\begin{equation} \label{eq:angular_cov}
\mathbf{\Sigma}_{\mathbf{\Omega}} = 
\begin{bmatrix}
\sigma_\theta^2 & 0 \\
0 & \sigma_\phi^2
\end{bmatrix}
\end{equation}
where $\sigma_\theta$ and $\sigma_\phi$ are the standard deviations of the measurement noise for azimuth and elevation, respectively.

\begin{figure}[t]
    \centering
    \includegraphics[width=0.95\columnwidth]{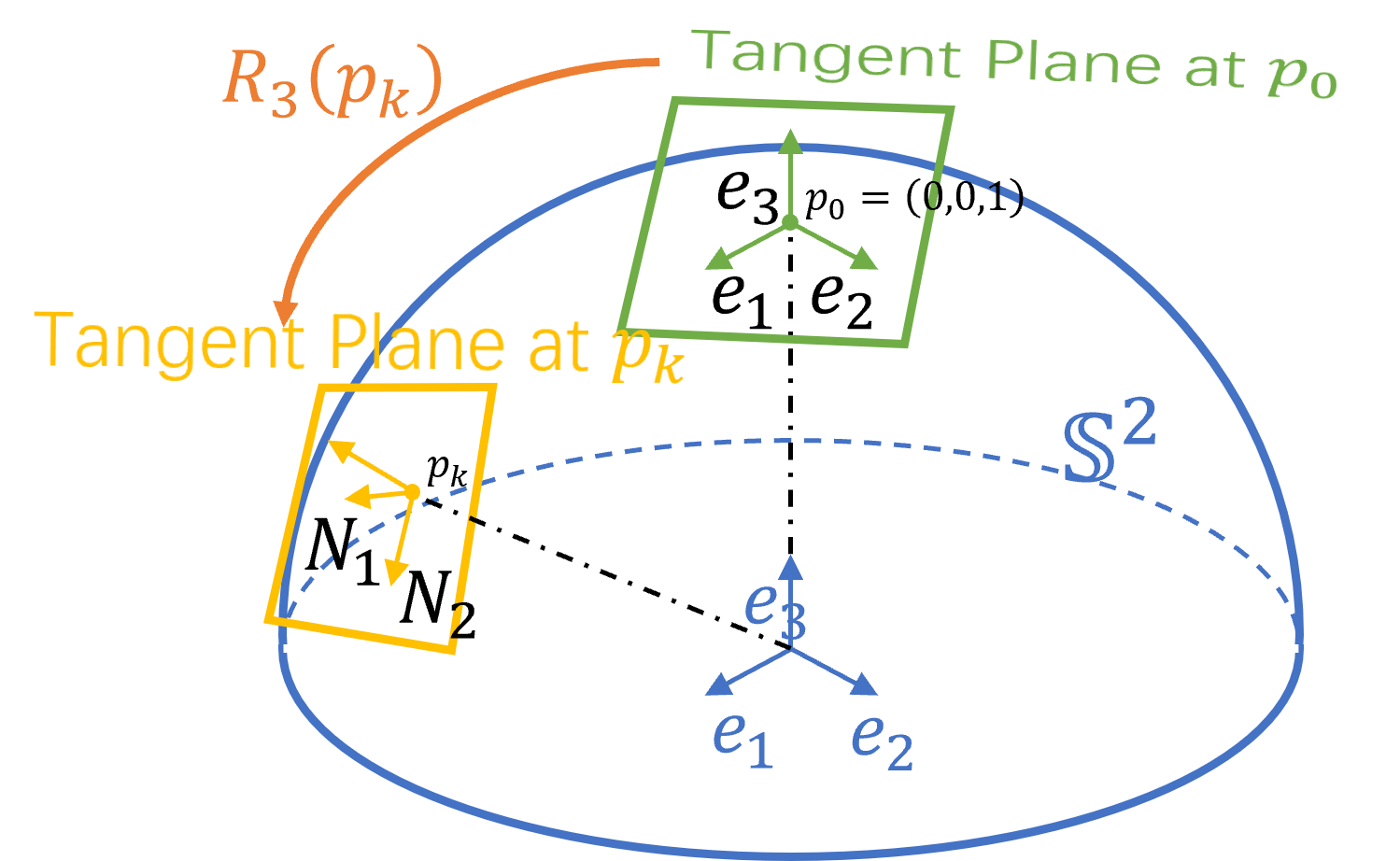}
    \caption{To obtain the orthonormal basis $(\mathbf{N}_1,\mathbf{N}_2)$ in the angent plane of $p_k \in \mathbb{S}^2$. $R_3(p_k)$ is defined by rotating $e_3$ to $p_k$, $\mathbf{N}_1$and $\mathbf{N}_2$ are obtained through rotating $e_1$ and $e_2$ by $R_3(p_k)$ respectively.}
    \label{fig:manifold}
\end{figure}

With the utilization of the $\boxplus$ operation in $\mathbb{S}^2$~\cite{he2021kalman}, we can build the connection between the ground-truth bearing direction $\mathbf{\Omega}^{gt}_k$ and its corresponding measurement $\mathbf{\Omega}_k$ as follows:
\begin{equation} \label{angle_noise}
\mathbf{\Omega}^{gt}_k=\mathbf{\Omega}_k\boxplus_{\mathbb{S}^2}\delta_{\mathbf{\Omega}_k}\triangleq e^{\left(\mathbf{N}(\mathbf{\Omega}_k) \delta_{\mathbf{\Omega}_k}\right)^\wedge}\mathbf{\Omega}_k
\end{equation}
in which the term $\mathbf{N}(\mathbf{\Omega}_k) = [\begin{matrix} \mathbf{N}_1 & \mathbf{N}_2 \end{matrix}] \in \mathbb{R}^{3\times2}$ represents an orthonormal basis, denoted as $\mathbf{N}(\mathbf{\Omega}_k)$, of the tangent plane at $\mathbf{\Omega}_k \in \mathbb{S}^2$. This can be visualized in Figure \ref{fig:manifold}. In addition, the notation $\left(\cdot\right)^\wedge$ refers to the skew-symmetric matrix that maps the cross product. The $\boxplus_{\mathbb{S}^2}$ operation involves rotating the unit vector $\mathbf{\Omega}_k$ around the axis $\delta_{\mathbf{\Omega}_k}$ within the tangent plane at $\mathbf{\Omega}_k$. This operation ensures that the resulting vector remains on the surface of $\mathbb{S}^2$.

Given a radar point $\prescript{R}{}{p_k}$ in polar coordinate, the ground-truth location of $\prescript{R}{}{p^{gt}_k}$ is formulated by combining the Equation~\eqref{r_noise} and \eqref{angle_noise}:
\begin{equation} \label{eq:pointwithnoise}
\begin{aligned}
\prescript{R}{}{p^{gt}_k} 
&= r^{gt}_k\mathbf{\Omega}^{gt}_k \\
&=(r_k+\delta_{r_k})(\mathbf{\Omega}_k\boxplus_{\mathbb{S}^2}\delta_{\mathbf{\Omega}_k}) \\
&\approx \underbrace{r_k\mathbf{\Omega}_k}_{\prescript{R}{}{p_k}} + \underbrace{\mathbf{\Omega}_k\delta_{r_k} - r_k\left(\mathbf{\Omega}_k\right)^\wedge\mathbf{N}(\mathbf{\Omega}_k)\delta_{\mathbf{\Omega}_k}}_{\prescript{R}{}{\mathbf{n}_{k}}}
\end{aligned}
\end{equation}
in which the term $\prescript{R}{}{\mathbf{n}_{k}}$ represents the noise of this radar point measurement: 
\begin{equation}
\prescript{R}{}{\mathbf{n}_{k}} = 
\underbrace{\left[\begin{matrix} \mathbf{\Omega}_k & -r_k\left(\mathbf{\Omega}_k\right)^\wedge\mathbf{N}(\mathbf{\Omega}_k)\end{matrix}\right]}_{\mathbf{A}_k} \left[ \begin{matrix} \delta_{r_k} \\  \mathbf{\delta_\Omega}_k \end{matrix} \right] \sim \mathcal{N}\left(\mathbf{0}, \prescript{R}{}{\mathbf{\Sigma}_k}\right)
\end{equation}
and the covariance $\prescript{R}{}{\mathbf{\Sigma}_k}$ can be written as follows:
\begin{equation}
\prescript{R}{}{\mathbf{\Sigma}_k} = \mathbf{A}_k \left[\begin{matrix} \sigma^2_r & \mathbf{0}_{1\times2} \\ \mathbf{0}_{2\times1} & \mathbf{\Sigma}_{\mathbf{\Omega}} \end{matrix}\right]\mathbf{A}^T_k
\label{eq:measurement_noise}
\end{equation}
\par To this end, we have modeled the radar measurement and its noise on the range and bearing direction, i.e., the uncertainty modeling of one radar point. In the following sections, we omit the frame $R$ in the representations that are related to the radar point.

One might argue that radar sensing provides not only range and angles but also Doppler velocities. However, Doppler velocity measurements usually contain non-Gaussian noise~\cite{kramer2020radar}, and therefore, modeling noise as a Gaussian white noise could yield inaccurate results. Additionally, relevant documentation indicates that Doppler velocity measurements are relatively precise, leading to our decision to omit Doppler velocity noise in this study. 


\subsection{Uncertainty-aware Residual on Doppler Velocity} 
\label{sec:DopplerConstraint}

Fusing Doppler velocity measurements is indispensable when the device works at a relatively high speed since it could reduce the error of IMU propagation. The Doppler velocity measurements reported by the radar should be the same as the projection of relative velocity between the radar and the object onto the direction vector of detection in the ideal scenario. This gives us the residual term $\mathbf{r}_{D}(\cdot)$ by fusing the Doppler velocity measurements.
\begin{equation} \label{eq:dopplerResidual}
\begin{aligned}
    \mathbf{r}_{D}(\mathbf{x}_{i},p_{k}) = \frac{{p_{k}}^{T}}{\lVert p_{k} \rVert} \cdot \mathbf{R}_{E}^{T} ({\mathbf{R}_{i}}^{T} \mathbf{v}_{i} + (\hat{\omega}_{i} - {\mathbf{b}_{g}}_{t})^{\wedge}\mathbf{t}_{E}) - {v_{d}}_{k}
\end{aligned}
\end{equation} 
where $p_{k}$ is the radar measurement in the $i$-th radar scan under a Cartesian coordinate system with the radar sensor as its origin point; ${v_{d}}_{k}$ is the Doppler velocity of $p_{k}$; $\hat{\omega}_{i}$ is the angular velocity measured by the IMU; $\mathbf{R}_{E}$ and $\mathbf{t}_{E}$ denote the extrinsic rotation and translation from the radar to the IMU, respectively. Note that the tangential velocity $(\hat{\omega}_{i} - {\mathbf{b}_{g}}_{t})^{\wedge}\mathbf{t}_{E}$ is considered in the residual formulation.


Directly fusing the residuals in Equation~\eqref{eq:OptProblem} without weighting will result in a deviation of the motion estimation, as different residuals ``contribute'' differently to the back-end estimation. Setting pre-defined weights may improve the results to some extent, however, we argue that the significance of each point measurement differs due to the physical properties of radar sensing. As described in Equation~\eqref{eq:pointwithnoise}, the uncertainty of a closer point is generally lower than that of a point located farther away from the radar sensor. The visualization of uncertainty-coupled radar points is presented in Figure~\ref{fig:points_with_cov}. Considering this, we utilize such modeled uncertainty to assign weights to the residuals based on their associated uncertainty levels. This approach allows for a more accurate fusion of the residuals and mitigates the impact of less reliable measurements on the back-end estimation.


Specifically, we integrate the noise term $\mathbf{n}_{k}$ into the Equation~\eqref{eq:dopplerResidual}, resulting in the following formulation using Taylor expansion:
\begin{equation}
\begin{aligned}
    0 &= \underbrace{\frac{({p_{k} + \mathbf{n}_{k}})^{T}}{\lVert p_{k} + \mathbf{n}_{k} \rVert}}_{\mathbf{d(N_k)}\in\mathbb{R}^{1\times3}} \cdot \underbrace{\mathbf{R}_{E}^{T} ({\mathbf{R}_{i}}^{T} \mathbf{v}_{i} + (\hat{\omega}_{i} - {\mathbf{b}_{g}}_{t})^{\wedge}\mathbf{t}_{E})}_{\mathbf{K}\in\mathbb{R}^{3\times1}} - {v_{d}}_{k} \\
    &\approx \mathbf{r}_{D}(\mathbf{x}_{i},p_{k})+\mathbf{J}_{\mathbf{n}_k}\mathbf{n}_k
\end{aligned}
\end{equation} 
where $\mathbf{r}_{D}(\cdot)$ is the residual of Doppler velocity, $\mathbf{n}_k$ is the noise of radar point measurement. The Jacobian $\mathbf{J}_{\mathbf{n}_k}$ can be represented as follows:
\begin{equation}
    \mathbf{J}_{\mathbf{n}_k}= \frac{\partial \mathbf{d}}{\partial \mathbf{n}_k}\cdot\mathbf{K}
\end{equation} 
\par Consequently, the covariance of the residual term is determined by employing the derivation presented in Equation~\eqref{eq:measurement_noise}:
\begin{equation} \label{eq:vel_residual_cov}
    \mathbf{\Sigma}_{\mathbf{r}_{D}(\mathbf{x}_{i},p_{k})} = \mathbf{J}_{\mathbf{n}_k} \mathbf{\Sigma}_k \ \mathbf{J}_{\mathbf{n}_k}^T
\end{equation}
\par Note that this covariance indicates the ``confidence'' of one radar point regarding the residual term $\mathbf{r}_{D}(\cdot)$. For instance, closer radar points are with higher confidence, leading to larger Doppler velocity-aided residuals in the optimization problem, thus making the state estimation more reasonable and accurate.


\subsection{Probability-guided Point Matching} 
\label{sec:ProbMatching}

\begin{figure*}[t]
    \centering
    \subfigure[The 3-$\sigma$ ellipses with polar radar measurement model]{
    \includegraphics[width=0.35\textwidth]{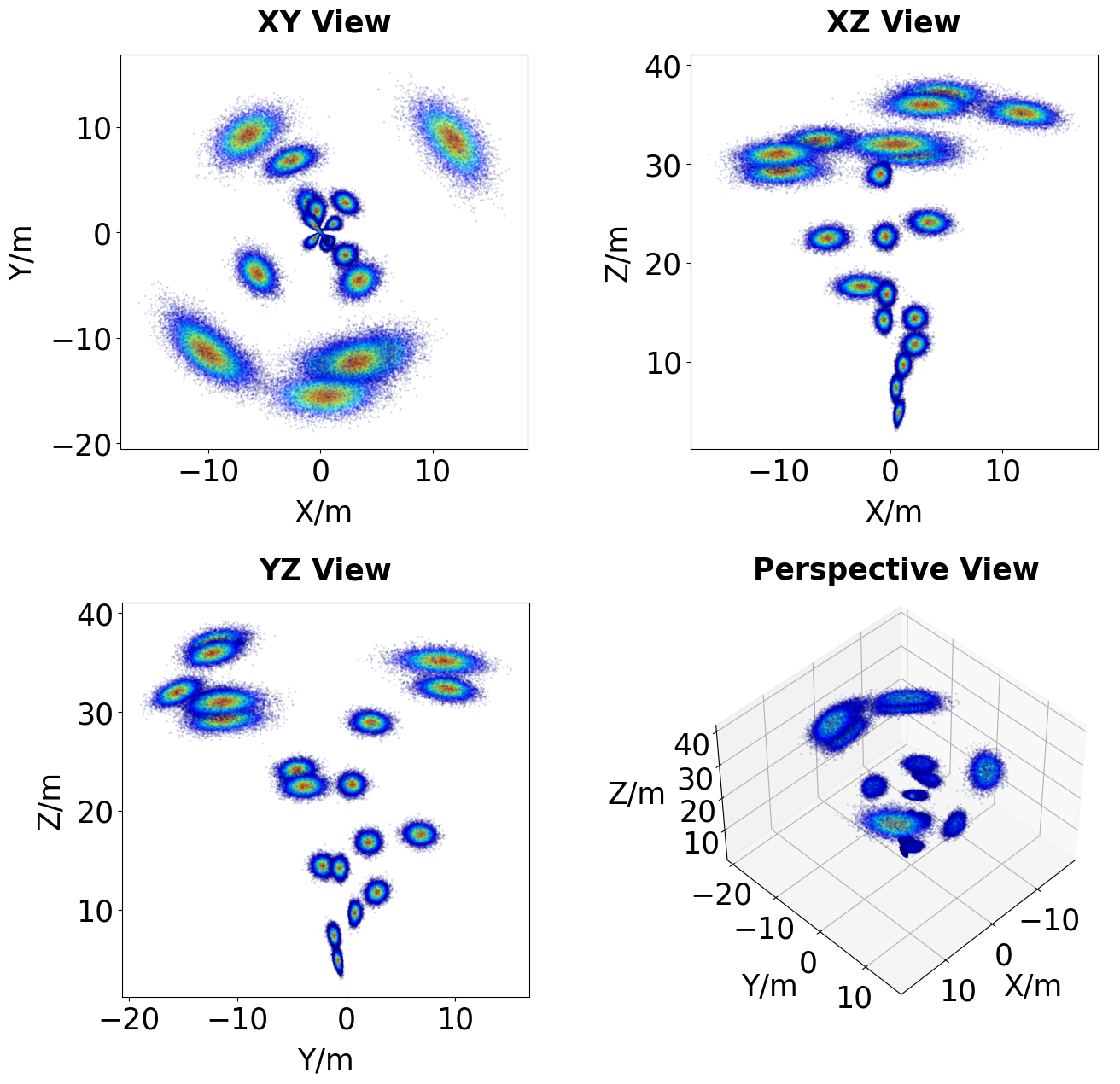}}
    \hfill
    \subfigure[The 3-$\sigma$ ellipses with polar and cartesian measurement models in XZ view
    ]{
    \includegraphics[width=0.50\textwidth]{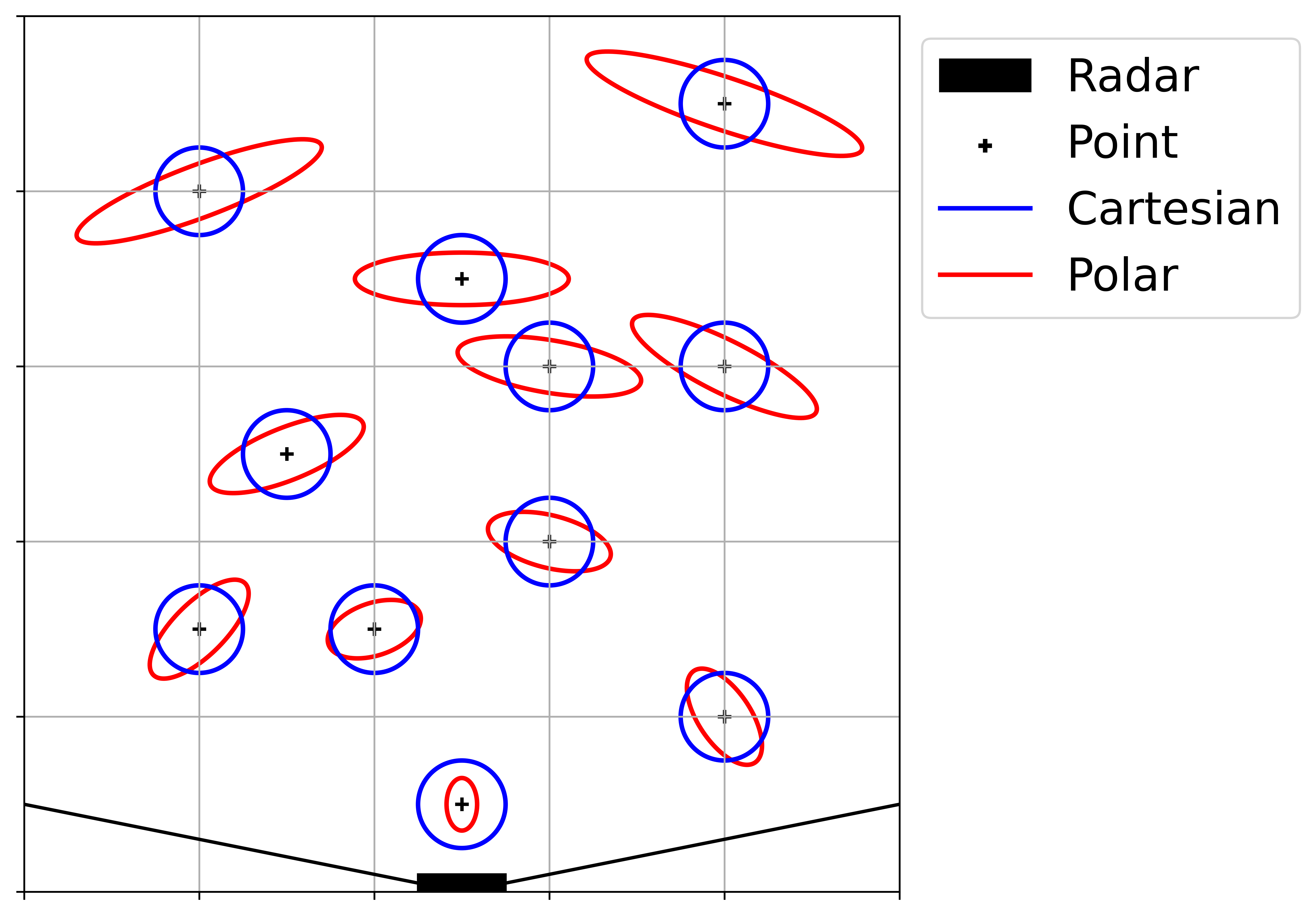}}
    \caption{(a) Visualization of 3-$\sigma$ ellipses with the proposed polar measurement model, and (0, 0, 0) is the coordinate of the radar sensor. The colors represent the probability density. (b) We also present the 3-$\sigma$ noise ellipses with cartesian radar measurement models.}
    \label{fig:points_with_cov}
\end{figure*}

Data association, or point matching, plays a pivotal role in point-based scan matching for robotic pose estimation~\cite{pomerleau2015review}. Prior research has demonstrated the feasibility of tracking radar points across successive radar scans~\cite{4DRadarSLAM}. In our previous study~\cite{Huang2024Less}, we employed a nearest neighbor-based search within Euclidean space. However, due to the inherent uncertainties in radar point measurements, the proximate points determined by this method do not always correspond to the most probable matches in a statistical sense. As delineated in Section~\ref{sec:NoiseModel} and Figure~\ref{fig:points_with_cov}, the probabilistic distribution of radar points within Euclidean space assumes an ellipsoidal shape due to the uncertainty model of the measurements. Thus, a shorter Euclidean distance to a central point does not necessarily indicate a higher probability of the point's presence at that location. This discrepancy suggests that employing a nearest-neighbor search in Euclidean space might lead to mismatches for state estimation.

To address this issue, we propose a probability-guided matching approach incorporating the uncertainty model outlined in Section~\ref{sec:NoiseModel}. This approach involves calculating the probability of a successful match for each point pair, thereby refining the data association step, i.e., the point-to-point matching process. This strategy aims to mitigate the limitations of traditional nearest-neighbor searches by incorporating the probabilistic distribution of modeled uncertainty into the data association step.

\begin{figure}[t]
    \centering
    \includegraphics[width=0.75\columnwidth]{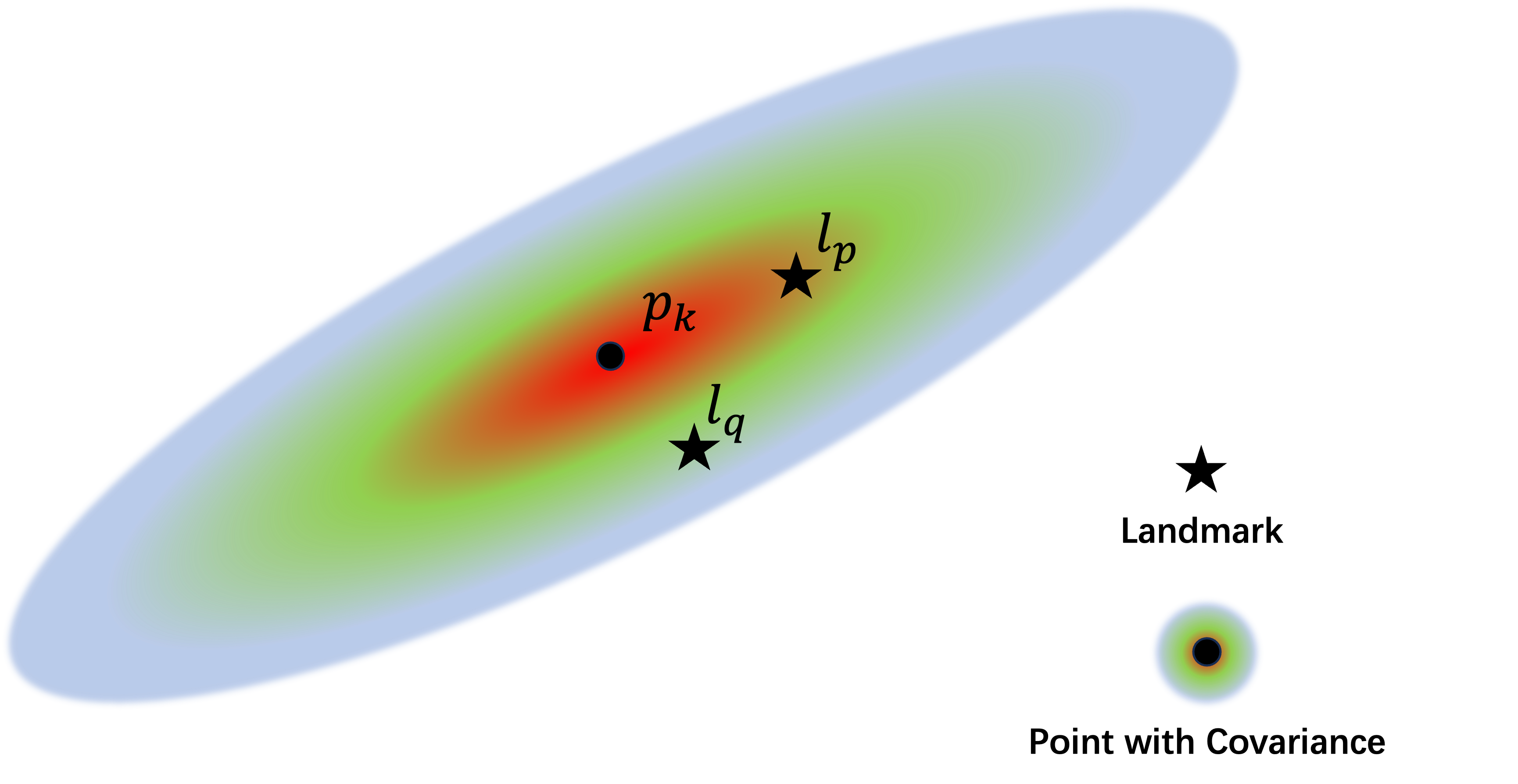}
    \caption{Given a radar point $p_k$, the matched map landmark is $l_p$ using our probability-guided point matching approach though $l_q$ is closer in Euclidean space. The colors represent the probability density.}
    \label{fig:match}
\end{figure}


Specifically, for a radar point $p_k$ viewed in polar coordinate attached in current radar pose, a number map landmarks $\mathbf{l}_m \in \mathbf{X}$ could be regarded as candidates. 
In the matching process, we compute the probability density $P_k(r, \theta, \phi)$ of each candidate landmark around the radar point $p_k$, as illustrated in Figure~\ref{fig:match}. In the proposed uncertainty model, $r, \theta, \phi$ are assumed to be mutually independent Gaussian random variables. Thus the matched probability of one landmark $l_q \in \mathbf{l}_m$ is $P_k$ and can be derived as:
\begin{equation} \label{eq:point_pdf}
\begin{aligned}
P_k(l_q)=
\frac{1}{\sqrt{(2 \pi)^3|{\mathbf{\Sigma}_k}|}} e^{-\frac{1}{2}(l_q-p_k)^{\mathrm{T}} {\mathbf{\Sigma}_k}^{-1}(l_q-p_k)}
\end{aligned}
\end{equation}
then the matched landmark that we are looking for is the one with the highest probability, stated as follows:
\begin{equation} \label{eq:matching_point}
\begin{aligned}
p_{\text{paired}} = \mathop{\arg\max}_{\mathbf{l}_m} P_k(l_q)
\end{aligned}
\end{equation}
\par We will also verify whether $p_{\text{paired}}$ is within the expected range using the 3-sigma rule. Points that do not fall within this range are considered outliers and removed. With this criterion and the probability-based matching, point correspondences are established in sequential frames, enabling the following point matching-based residual function.


One might argue that the uncertainty of map landmarks should be considered. However, the covariance of map points is generally based on the voxelization and appearance of dense map points~\cite{jiang2022lidar,yuan2022efficient}. Radar map points are much sparser and noisier than LiDAR points, making it difficult to estimate the covariance of map landmarks. Building multiple-to-multiple matching is also inapplicable with such sparse points. Thus, in this study, we only model the uncertainty of observed radar points for one-to-one correspondence estimation.

\subsection{Uncertainty-aware Residual on Point Matching} 
\label{sec:PointConstraint}

Finally, with the estimated point matching in Section~\ref{sec:ProbMatching}, we formulate the point-to-point residual term $\mathbf{r}_{P}(\cdot)$ as follows:
\begin{equation} \label{eq:pointResidual}
\begin{aligned}
    \mathbf{r}_{P}(\mathbf{x}_{i},\mathbf{l}_{k},p_{k}) = \mathbf{l}_{k} - (\mathbf{R}_{i}(\mathbf{R}_{E} p_{k} + \mathbf{t}_{E}) + \mathbf{p}_{i})
\end{aligned}
\end{equation}
where $\mathbf{l}_{k}$ is the position of the matched landmark; $p_{k}$ is the observed point of $\mathbf{l}_{k}$ in the $i$-th scan. The residual is formulated by considering the distance between the landmark and the measurement point corresponding to it.

The inclusion of the noise term $\mathbf{n}_{k}$ results in the formulation as follows:
\begin{equation}
\begin{aligned}
    0 &= \mathbf{l}_{k} - (\mathbf{R}_{i}(\mathbf{R}_{E} (p_{k} + \mathbf{n}_k) + \mathbf{t}_{E}) + \mathbf{p}_{i}) \\
    &\approx \mathbf{r}_{P}(\mathbf{x}_{i},\mathbf{l}_{k},p_{k})+\mathbf{J}_{\mathbf{n}_k}\mathbf{n}_k
\end{aligned}
\end{equation}
in which the term $\mathbf{r}_P(\cdot)$ refers to the residual on point-to-point matching, while $\mathbf{n}_k$ represents the noise of one radar point measurement $p_k$. The Jacobian term $\mathbf{J}_{\mathbf{n}_k}$ can be derived as follows:
\begin{equation}
\mathbf{J}_{\mathbf{n}_k}=\mathbf{R}_{i}\mathbf{R}_{E}
\end{equation}
then, the covariance to adjust the residual is:
\begin{equation}
 \mathbf{\Sigma}_{\mathbf{r}_{P}(\mathbf{x}_{i},\mathbf{l}_{k},p_{k})} = \mathbf{J}_{\mathbf{n}_k} \mathbf{\Sigma}_k \ \mathbf{J}_{\mathbf{n}_k}^T
\end{equation}




Similar to the aforementioned residual $\mathbf{r}_D(\cdot)$, each radar point is with different noise, thus the proposed uncertainty model results in the adjustment of weights that are assigned to the residual $\mathbf{r}_P(\cdot)$. 

We have proposed our uncertainty model and incorporation schemes that could answer the two questions in Section~\ref{sec:Introduction}: \textit{How do we model the uncertainty of radar measurements?} (Section~\ref{sec:NoiseModel}) and \textit{How do we enhance the performance of radar SLAM with the uncertainty modeling?} (Section~\ref{sec:DopplerConstraint}, \ref{sec:ProbMatching} and \ref{sec:PointConstraint}). We conduct real-world tests for experimental validation in the following experimental sections.


\section{Experiments} \label{sec:Experiments}




\subsection{Experimental Setup}
\label{sec:setup}

\begin{figure}[t]
    \centering
    \includegraphics[width=0.75\columnwidth]{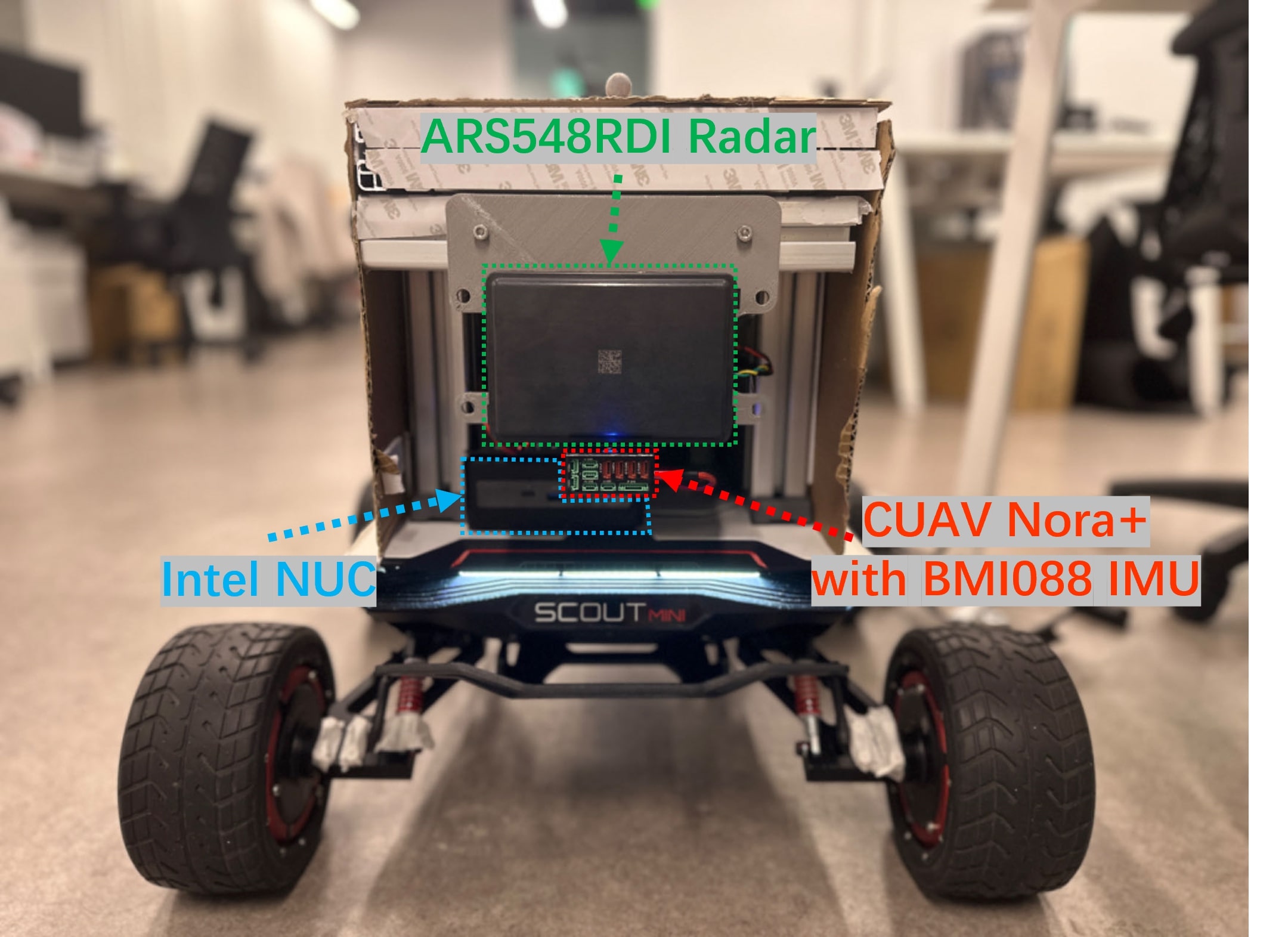}
    \caption{Our customized vehicle for experimental validation. The vehicle is equipped with a radar sensor, an IMU and a computing platform. We validate our proposed approaches within a controlled room outfitted with a motion capture system.}
    \label{fig:platform}
\end{figure}

Two different datasets are utilized: one is our self-collected dataset on a mobile platform, shown in Figure~\ref{fig:platform}, and the other is the public radar dataset ColoRadar~\cite{kramer2021coloradar}. Our platform consists of a 4D FMCW Radar ARS548RDI~\footnote{The document can be fount at \href{https://conti-engineering.com/wp-content/uploads/2023/01/RadarSensors\_ARS548RDI.pdf}{Continental ARS548RDI}.} manufactured by Continental and an IMU BMI088 manufactured by Bosch. We utilize a motion capture system to obtain precise ground truth poses for evaluation. This system allows us to collect three distinct sequences for evaluation, each presenting varying levels of difficulty: \textbf{Sequence 1} involved relatively low-speed movements, serving as a baseline for comparison; \textbf{Sequence 2} is collected at higher speeds, introducing additional challenges; and \textbf{Sequence 3} is the most challenging, characterized by high vehicle speeds.


The ColoRadar dataset is collected by a single-chip Texas Instruments radar sensor, which provides sparser points. We select three indoor sequences: ``12\_21\_2020\_arpg\_lab\_run0'', ``12\_21\_2020\_ec\_hallways\_run0'', ``2\_24\_2021\_aspen\_run0'' referred to as \textbf{ColoRadar 1}, \textbf{ColoRadar 4} and \textbf{ColoRadar 5}; one outdoor sequence ``2\_28\_2021\_outdoor\_run0'', denoted as \textbf{ColoRadar 2}; and one sequence collected in a narrow subterranean environment: ``2\_23\_2021\_edgar\_classroom\_run0'', denoted as \textbf{ColoRadar 3}. ColoRadar 5 is collected in a lab environment with a motion capture system providing ground truth. For other sequences, a high-precision and finely-tuned LiDAR-inertial odometry system is applied as the ground truth. All sequences are acquired using a hand-held platform with aggressive motion, posing significant challenges for motion estimation.

It is worth noting that the sensor characteristics of these two datasets are different. Thus, the covariance settings also vary accordingly. In our self-collected dataset, we set the 3-sigmas of Gaussians equal to the “largest detection errors” provided by the sensor datasheet, thus obtaining the values of $\sigma_r$ and $\mathbf{\Sigma}_{\mathbf{\Omega}}$. The ColoRadar dataset does not provide such parameters for uncertainty modeling. We fine-tune the covariance settings using the ColoRadar 1 sequence and then apply the same parameters to the other four sequences. Additionally, in our previous work~\cite{Huang2024Less}, we enhanced the radar-inertial odometry with radar cross-section information, which is a type of intensity of radar points. In this study, we disable all the physically-enhanced modules to test the proposed approaches clearly and thoroughly.


\subsection{Do We Need to Incorporate the Point Uncertainty?}

\begin{figure*}[t]
    \centering
    \includegraphics[width=0.95\textwidth]
    {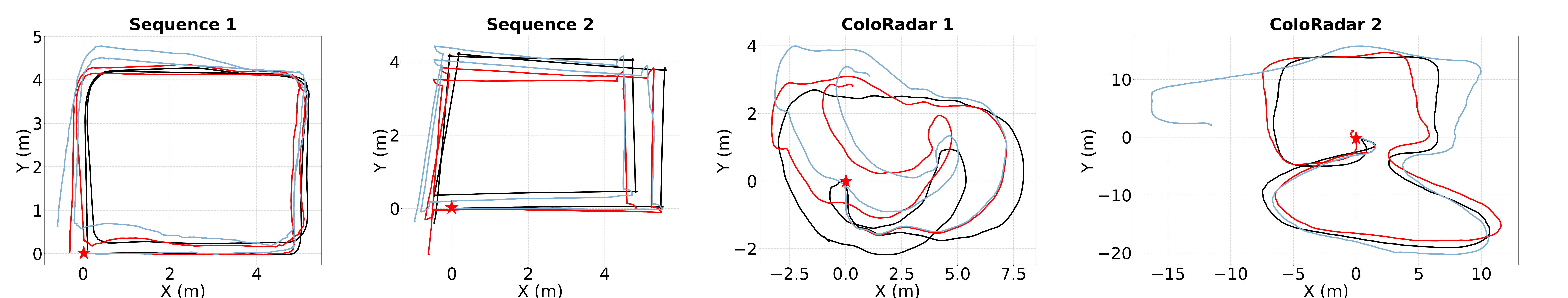}
    \caption{\textcolor{red}{{Red}} trajectory is the proposed full system. \textcolor{blue}{{Blue}} one is the system without uncertainty model. \textcolor{black}{{Black}} one is the trajectory with ground truth poses. The starting points are marked as{red star}. We present the results on four sequences in two different datasets.}
    \label{fig:UM_traj}
\end{figure*}

\begin{figure}[t]
    \centering
    \includegraphics[width=0.95\columnwidth]{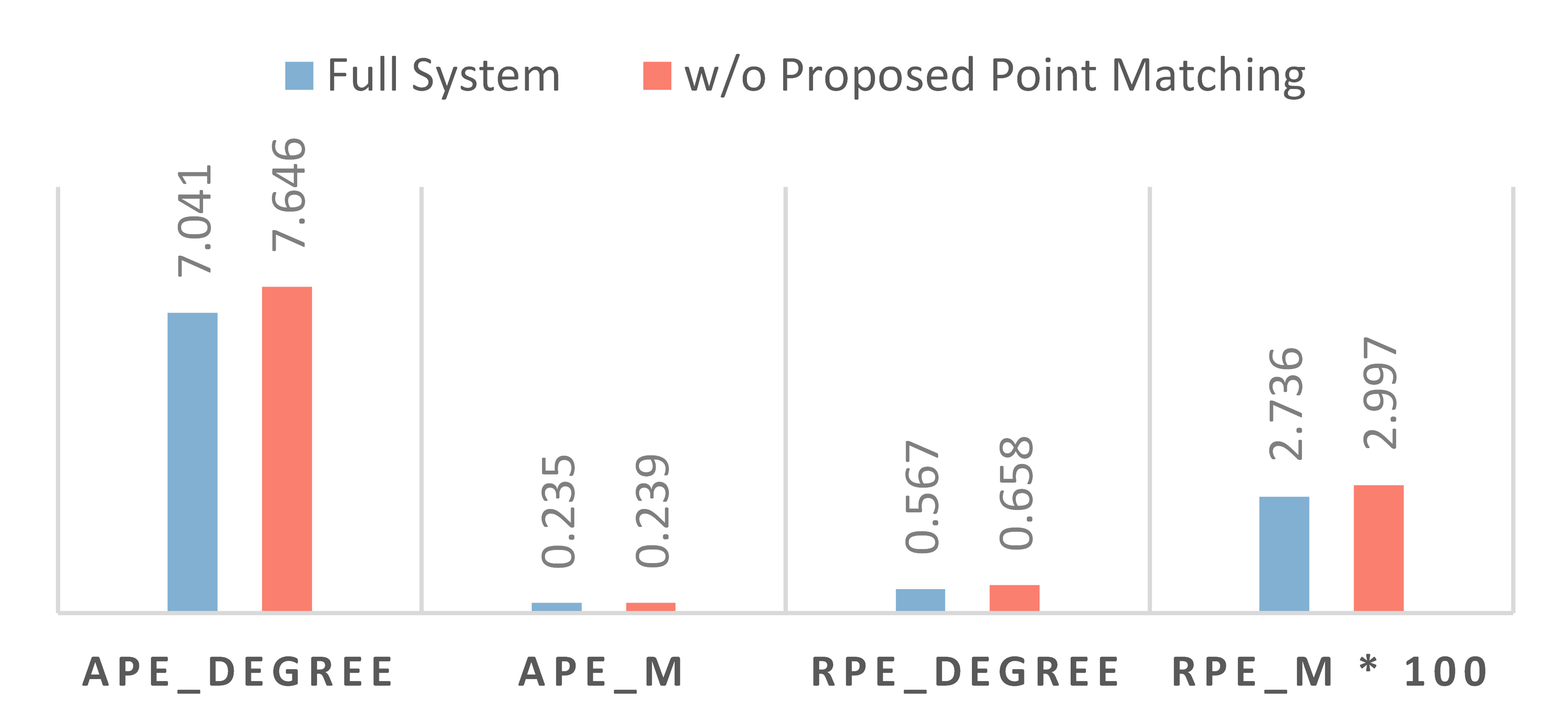}
    
    \caption{The comparison with the proposed probability-guided matching and the conventional Euclidean-based matching. The radar SLAM performs better if the modeled uncertainty is leveraged into the data association. }
    \label{fig:wo_UM_Match}
\end{figure}

We compare the full system with several uncertainty-eliminated versions, i.e., setting $\sigma_r$ or $\mathbf{\Sigma}_{\mathbf{\Omega}}$ to a very small value. Essentially, this will affect the weights assigned to the residuals in the optimization process, leading to different motion estimation results. To evaluate the ablated systems, we conducted tests using four sequences encompassing indoor and outdoor scenes with varying platform speeds. Table \ref{tab:noise_model_performance} and Figure \ref{fig:UM_traj} present the quantitative results and the estimated motion trajectories, respectively.

\begin{table}[t] 
\begin{center}   
\caption{Quantitative Results of Ablation Study\\ on Uncertainty Modeling (UM)}  \label{tab:noise_model_performance} 
\resizebox{\columnwidth}{!}{
\begin{tabular}{cccccc} 
    \hline
    \hline
        \multirow{2}*{\textbf{Sequence}} &\multirow{2}*{\textbf{method}} & \multicolumn{2}{c}{\textbf{APE RMSE}} & \multicolumn{2}{c}{\textbf{RPE RMSE}} \\
        \cline{3-4} \cline{5-6}
        & &Trans.(m) &Rot.($^{\circ}$) &Trans.(m) &Rot.($^{\circ}$) \\
    \hline
        \multirow{4}*{Sequence 1} 
        & w/o UM         & 0.505          & 11.485         & 0.032          & 1.141 \\ 
        & w/o Range UM   & 0.480          & 10.087         & 0.041          & 1.595 \\
        & w/o Agular UM  & 0.562          & 10.859         & 0.045          & 1.596 \\  
        & Ours           & \textbf{0.235} & \textbf{7.041} & \textbf{0.027} &\textbf{0.566}\\ 
    \hline
        \multirow{4}*{Sequence 2} 
        & w/o UM         & \textbf{0.333} & 10.293         & 0.059           & 1.101 \\
        & w/o Range UM   & 0.933          & 20.611         & 0.067           & 1.819 \\ 
        & w/o Agular UM  & 0.956          & 20.159         & 0.072           & 1.763 \\ 
        & Ours           & 0.586          & \textbf{5.415} & \textbf{0.051}  & \textbf{0.818}\\ 
    \hline
        \multirow{4}*{ColoRadar 1} 
        & w/o UM         & 6.187          & 22.750         & 0.045            & 1.067 \\ 
        & w/o Range UM   & 6.198          & 22.988         & 0.045            & 1.078 \\
        & w/o Agular UM  & 6.192          & 22.869         & 0.045            & 1.075 \\ 
        & Ours           & \textbf{6.101} & \textbf{12.640}& \textbf{0.044}   & \textbf{0.981}\\ 
    \hline
        \multirow{4}*{ColoRadar 2} 
        & w/o UM         & 9.466          & 52.867          & 0.041           & 2.418 \\ 
        & w/o Range UM   & 8.063          & 28.949          & 0.042           & 2.581 \\
        & w/o Agular UM  & 11.508         & 65.591          & 0.120           & 2.373 \\ 
        & Ours           & \textbf{7.469} & \textbf{11.983} & \textbf{0.038}  & \textbf{2.162}\\ 
    \hline
    \hline
\end{tabular}
}
\end{center}   
\end{table}

The experimental results indicate that incorporating uncertainty measurement in the system leads to superior performance in terms of most evaluation metrics across various sequences. Systems without modeled uncertainty or with partial implementation exhibit inferior performance in comparison. This finding highlights the necessity of the uncertainty model and also validates the effectiveness of the incorporation scheme proposed in this study.

Despite the ablation studies above, we also test the proposed data association approach individually. We compare our approach with the Euclidean distance-based nearest neighbor search, which is a widely used approach for range sensing-based SLAM systems, and is also applied in our previous work~\cite{Huang2024Less}. As shown in Figure~\ref{fig:wo_UM_Match}, our approach has an overall improvement compared to the Euclidean-based method, thus validating the effectiveness of the probability-guided point matching with the modeled uncertainty.

\subsection{Is the Covariance Setting Right?}

As mentioned in Section~\ref{sec:setup}, the covariance settings are directly based on the radar datasheet when using our self-collected dataset. In this section, we will examine the pivotal role of uncertainty parameters in the radar SLAM system by changing covariance settings accordingly. Theoretically, when the parameters are set to relatively large values, more radar points will be assigned smaller weights in the estimation process. Hence, IMU measurements will have a larger impact on motion estimation, even though radar measurements are inherently more accurate. This imbalance can lead to a biased state estimation. On the other hand, if the uncertainty parameters are set to smaller values, it will also lead to a biased estimation. Therefore, finding an appropriate balance and accurately calibrating the uncertainty parameters is crucial to ensure that radar measurements are properly weighted and contribute more reasonably to the radar SLAM system.

\begin{figure}[t]
    \centering
    \includegraphics[width=0.95\columnwidth]{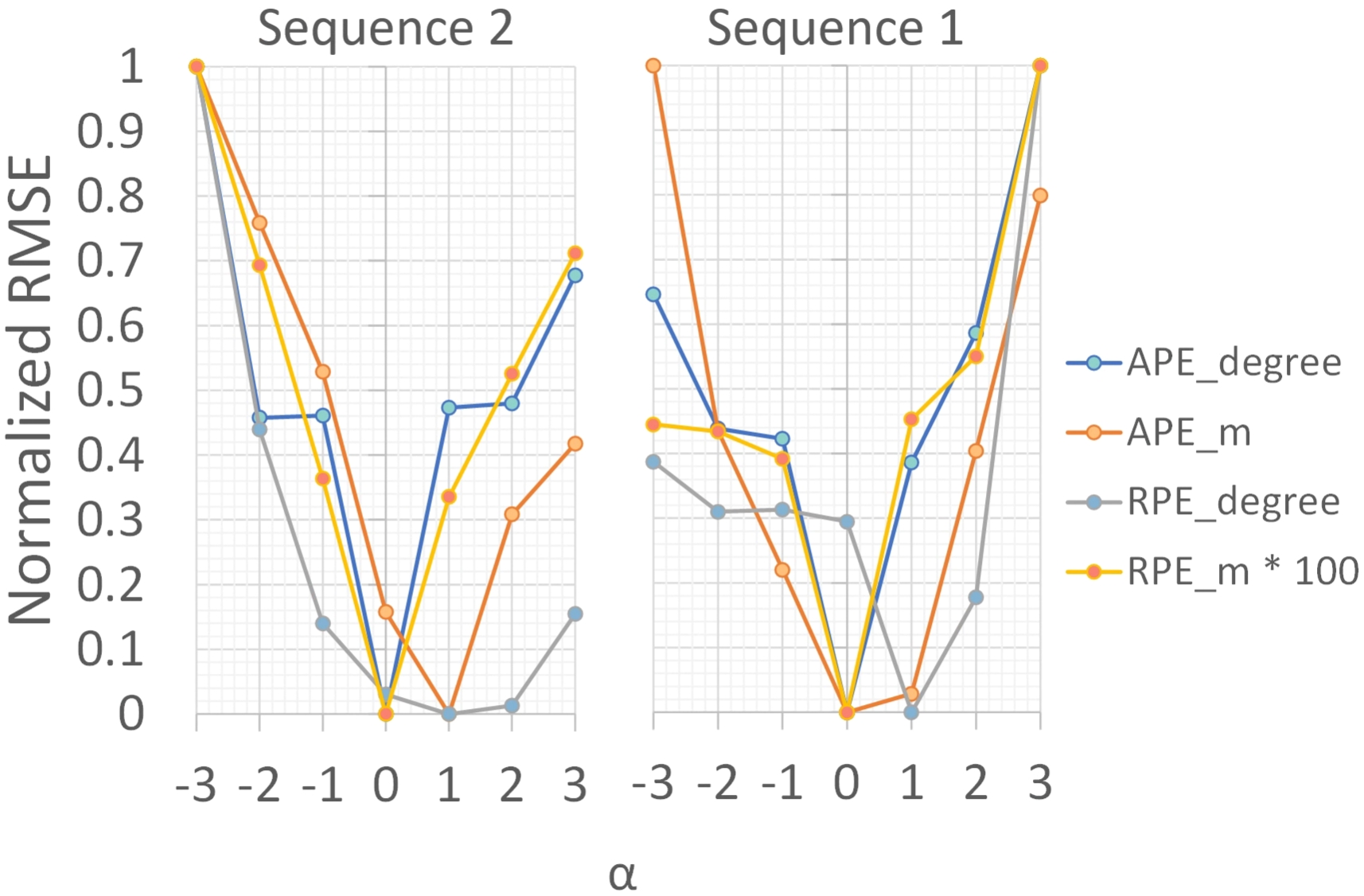}
    \caption{Normalized RMSE within the range of 0 to 1, and x-axis represents the coefficient $2^\alpha$ from $2^{-3}$ to $2^3$ shown in Table \ref{tab:uncertainty_parameter_analysis}.}
    \label{fig:uncertainty_parameter_analysis}
\end{figure}

To validate our analysis above, we conduct experiments on Sequence 1 and Sequence 2, and we apply the coefficients to the parameters of uncertainty models (covariances). Specifically, the parameters are changed to $\mathbf{\Sigma} = 2^{\alpha} {\mathbf{\Sigma}}$, where $\alpha$ denotes the coefficient ranging from -3 to 3 and $\mathbf{\Sigma}=(\sigma_r, \mathbf{\Sigma}_{\mathbf{\Omega}})$. The experimental results are summarized in Table~\ref{tab:uncertainty_parameter_analysis}. To provide a clearer understanding, we also present the results graphically in Figure~\ref{fig:uncertainty_parameter_analysis}: the normalized root mean square error (RMSE) ranges from 0 to 1, i.e., ranges from the best performance to the worst one.

\begin{table}[t] 
\begin{center}   
\caption{Quantitative Results of Different Sets of Modeled Uncertainty on Self-Collect Dataset}  \label{tab:uncertainty_parameter_analysis} 
\begin{tabular}{cccccc} 
    \hline
    \hline
        \multirow{2}{*}{\textbf{Sequence}} &\multirow{2}{*}{$\alpha$} & \multicolumn{2}{c}{\textbf{APE RMSE}} & \multicolumn{2}{c}{\textbf{RPE RMSE}} \\
        \cline{3-6} 
        & &Trans.(m) &Rot.($^{\circ}$) &Trans.(m) &Rot.($^{\circ}$) \\
    \hline
     \multirow{7}*{\makecell[c]{Sequence 1}}
        & -3  & 0.435  & 8.703  & 0.0300  & 0.577 \\ 
        & -2  & 0.322  & 8.169  & 0.0299  & 0.568 \\ 
        & -1  & 0.279  & 8.128  & 0.0297  & 0.568 \\ 
        & 0   & \textbf{0.235}  & \textbf{7.041}  & \textbf{0.0273}  & 0.566 \\ 
        & 1   & 0.241  & 8.034  & 0.0300  & \textbf{0.530} \\ 
        & 2   & 0.316  & 8.549  & 0.0306  & 0.552 \\ 
        & 3   & 0.395  & 9.613  & 0.0333  & 0.652 \\ 
    \hline
    \multirow{7}*{\makecell[c]{Sequence 2}}
        & -3  & 0.700  & 8.570  & 0.0625  & 1.495 \\ 
        & -2  & 0.668  & 6.859  & 0.0592  & 1.104 \\ 
        & -1  & 0.636  & 6.868  & 0.0556  & 0.895 \\ 
        & 0   & 0.586  & \textbf{5.415}  & \textbf{0.0517}  & 0.818 \\ 
        & 1   & \textbf{0.564}  & 6.908  & 0.0553  & \textbf{0.797} \\ 
        & 2   & 0.606  & 6.928  & 0.0574  & 0.806 \\ 
        & 3   & 0.621  & 7.552  & 0.0594  & 0.905 \\ 
    \hline
    \hline
\end{tabular}
\end{center}   
\end{table}

In Table~\ref{tab:uncertainty_parameter_analysis} and Figure~\ref{fig:uncertainty_parameter_analysis}, the overall performance drops when the covariances are either too small or too large. This could support our analysis of the uncertainty of radar measurements. Interestingly, we find that the parameters provided in the radar datasheet ($\alpha$=1) tend to yield the best performance. The best performance is achieved when $\alpha$ is in the range of $\left[ 1,1.5\right]$. We consider that this is mainly due to two reasons. First, the largest error provided by the datasheet is not equal to the 3-sigma in the Gaussian distribution. Further studies on radar sensing are required if we need to set the covariances precisely. Second, these datasheet parameters may not be optimal due to variations among individual radar sensors. It is necessary to calibrate the covariance settings to achieve better performance. In summary, the specific parameters (largest detection errors) from the datasheet are critical for modeling the uncertainty of radar points, and they could help enhance the performance using the proposed approaches of this study. Further studies are required to estimate such parameters precisely via deep analyses and real-world tests.


\begin{table}[t]    
\begin{center}   
\caption{Quantitative Results of Comparison Study}  \label{tab:performance_compare}
\begin{tabular}{cccccc}
\hline
\hline
    \multirow{2}*{\textbf{Sequence}} &\multirow{2}*{\textbf{Method}} & \multicolumn{2}{c}{\textbf{APE RMSE}} & \multicolumn{2}{c}{\textbf{RPE RMSE}} \\
    \cline{3-4} \cline{5-6}
    & &Trans.(m) &Rot.($^{\circ}$) &Trans.(m) &Rot.($^{\circ}$) \\
\hline
    \multirow{4}*{\makecell[c]{Sequence 1\\Dis. 35.3m}} 
    & EKF-RIO    & 0.800          & 15.338         & 0.114          & \textbf{0.499} \\ 
    & Ours-P2D       & 0.834          & 8.485         & 0.069          & 0.720 \\ 
    & Ours-D2M       & 1.194          & 21.457         & \textbf{0.027}          & 0.646 \\ 
    & Ours      & \textbf{0.235} & \textbf{7.041} & \textbf{0.027} & 0.566 \\ 
\hline
    \multirow{4}*{\makecell[c]{Sequence 2\\Dis. 38.6m}} 
    & EKF-RIO    & 2.805          & 8.835          & 0.392          & 0.866 \\ 
    & Ours-P2D      & 0.743          & 5.677          & 0.061 & \textbf{0.805} \\ 
    & Ours-D2M       & 87.189         & 8.345          & 2.148          & 0.806 \\ 
    & Ours      & \textbf{0.586} & \textbf{5.415} & \textbf{0.051} & 0.818 \\ 
\hline
    \multirow{4}*{\makecell[c]{Sequence 3\\Dis. 47.8m}} 
    & EKF-RIO    & 68.960         & 66.228         & 2.195          & 2.111 \\ 
    & Ours-P2D       & 2.021          & \textbf{12.729}         & \textbf{0.052}          & 2.241 \\ 
    & Ours-D2M       & 3.277          & 21.091         & 0.273          & 2.411 \\ 
    & Ours      & \textbf{1.791} & 14.805 & \textbf{0.052} & \textbf{2.105}  \\ 
\hline
    \multirow{4}*{\makecell[c]{ColoRadar 1\\Dis. 101.0m}}
    & EKF-RIO    &\textbf{5.348}  &24.416          &0.054           &6.044  \\ 
    & Ours-P2D       &5.964           &13.030          &0.067           &0.77  \\ 
    & Ours-D2M       &5.993           &13.605          &0.066           &\textbf{0.743}  \\ 
    & Ours      &6.101           &\textbf{12.640} &\textbf{0.044}  &0.981  \\ 
\hline
    \multirow{4}*{\makecell[c]{ColoRadar 2\\Dis. 116.4m}}
    & EKF-RIO    &10.692          &14.764          &\textbf{0.047}            &\textbf{0.497} \\ 
    & Ours-P2D       &11.159          &20.024          &0.054               &2.098 \\
    & Ours-D2M       &10.642          &15.623 &0.054               &2.079 \\
    & Ours      &\textbf{5.993}  &\textbf{13.605}          &0.066      &0.7 \\
\hline
    \multirow{4}*{\makecell[c]{ColoRadar 3\\Dis. 188.4m}}
    & EKF-RIO    &10.995          &37.731          &0.052            &2.589 \\ 
    & Ours-P2D       &8.383          &\textbf{16.497}          &0.052               &2.578 \\
    & Ours-D2M       &14.058          &35.566        &0.094               &2.635 \\
    & Ours      &\textbf{8.340}  &29.367          &\textbf{0.040}      &\textbf{0.553} \\
\hline
    \multirow{4}*{\makecell[c]{ColoRadar 4\\Dis. 112.0m}}
    & EKF-RIO    &9.352          &18.156          &0.086            &\textbf{0.516} \\ 
    & Ours-P2D       &7.628          &24.341          &0.075               &2.382 \\
    & Ours-D2M       &12.652          &29.135 &0.311               &2.241 \\
    & Ours      &\textbf{5.223}  &\textbf{16.070}          &\textbf{0.049}      &0.675 \\
\hline
    \multirow{4}*{\makecell[c]{ColoRadar 5\\Dis. 44.7m}}
    & EKF-RIO    &4.337          &52.547          &0.044               &0.668 \\ 
    & Ours-P2D       &4.767           &44.934          &0.044               &1.404 \\
    & Ours-D2M       &4.311           &45.821          &0.055               &\textbf{0.60} \\
    & Ours      &\textbf{3.820}  &\textbf{30.905} &\textbf{0.036}      &1.146 \\
\hline
\hline
\end{tabular}
\end{center}
\end{table}

\subsection{Performance of Uncertainty-aware RIO}


\indent This section evaluates the uncertainty-aware RIO system with comparisons. The comparisons include EKF-RIO~\cite{doer2020ekf}, and two distribution-based measurement models: Point-to-Distribution (P2D), and Distribution-to-Multi-Distribution (D2M)~\cite{zhuang20234d}. EKF-RIO is an open-source radar-inertial odometry system. Both P2D and D2M share similarities with our approach but differ in the correspondence estiamtion and residuals. P2D applies conventional laser-scan matching techniques, while D2M constructs a distribution for each current point by identifying neighboring points. We implement P2D and D2M by ourselves and integrate them into the proposed RIO framework, referring to Ours-P2D and Ours-D2M, respectively. Furthermore, we enhance their performance by building submaps, following the guidelines outlined in~\cite{zhuang20234d}.

\indent We employ evo~\cite{grupp2017evo} to evaluate performance in terms of relative pose error (RPE) and absolute pose error (APE). The calculation of RPE depends on the data interval, which can be determined either by sampling fixed-interval point pairs along the trajectory or by forming data pairs using consecutive frames. In this study, we use the ``per-frame'' for RPE calculation, which is the default setting of evo. The quantitative results are summarized in Table~\ref{tab:performance_compare}. Our method performs better on the RMSE metrics in the self-collected sequences. Specifically, in terms of APE RMSE, our method outperforms the other methods. In the ColoRadar dataset, which is collected using a single-chip radar sensor, the presence of radar measurement noise presents challenges for all methods. EKF-RIO, in this case, exhibits instability due to discarding too many radar measurements. It occasionally provides good results, but it could also introduce drift in both translation and rotation. Ours-P2D and Ours-D2M consider the geometry information surrounding radar points for matching. Due to the noise and sparsity of radar data, the distribution-based estimation will result in occasional larger errors compared to our point-only matching method. It is worth noting that the distribution-based 4D iRIOM~\cite{zhuang20234d} performs much better than our system and is competitive to LiDAR-based Fast-LIO~\cite{xu2021fast}. This is due to many factors, such as the loop closing, the robust estimation, and the system implementation. This study focuses on the feasibility and effectiveness of incorporating point uncertainty into radar state estimation. To achieve high accuracy estimation, we recommend users to follow the system design and implementation in the 4D iRIOM~\cite{zhuang20234d}. 

In the recent work by Kubelka \textit{et al.}~\cite{kubelka2023we}, the experimental results show that the odometry from Doppler and IMU data even performs better than scan matching-only methods. We consider this to depend on several factors, such as sensor characteristics. In this study, we fuse the Doppler velocity, IMU data, and point matching together to build a radar-inertial odometry system and demonstrate that point matching plays a vital role in the sensor fusion framework. This demonstration can be found in the experimental results of this paper and of our previous work~\cite{Huang2024Less}.



\section{Conclusion} \label{sec:Conclusion}

In this study, we model the uncertainty of radar points and incorporate it into the RIO system. Radar has become a new wave in recent years and has been applied to many robotic applications~\cite{harlow2023new}. Several potential problems are still not solved well. In the future, we aim to explore promising directions such as multi-modal fusion and robust estimation. Additionally, the development of a specialized dataset is crucial for the comprehensive evaluation of per-point uncertainty.



\bibliographystyle{IEEEtran}
\bibliography{root}

\begin{thebibliography}{10}
\providecommand{\url}[1]{#1}
\csname url@samestyle\endcsname
\providecommand{\newblock}{\relax}
\providecommand{\bibinfo}[2]{#2}
\providecommand{\BIBentrySTDinterwordspacing}{\spaceskip=0pt\relax}
\providecommand{\BIBentryALTinterwordstretchfactor}{4}
\providecommand{\BIBentryALTinterwordspacing}{\spaceskip=\fontdimen2\font plus
\BIBentryALTinterwordstretchfactor\fontdimen3\font minus \fontdimen4\font\relax}
\providecommand{\BIBforeignlanguage}[2]{{%
\expandafter\ifx\csname l@#1\endcsname\relax
\typeout{** WARNING: IEEEtran.bst: No hyphenation pattern has been}%
\typeout{** loaded for the language `#1'. Using the pattern for}%
\typeout{** the default language instead.}%
\else
\language=\csname l@#1\endcsname
\fi
#2}}
\providecommand{\BIBdecl}{\relax}
\BIBdecl

\bibitem{hong2022radarslam}
Z.~Hong, Y.~Petillot, A.~Wallace, and S.~Wang, ``Radarslam: A robust simultaneous localization and mapping system for all weather conditions,'' \emph{The International Journal of Robotics Research}, vol.~41, no.~5, pp. 519--542, 2022.

\bibitem{yin2021rall}
H.~Yin, R.~Chen, Y.~Wang, and R.~Xiong, ``Rall: end-to-end radar localization on lidar map using differentiable measurement model,'' \emph{IEEE Transactions on Intelligent Transportation Systems}, vol.~23, no.~7, pp. 6737--6750, 2021.

\bibitem{kramer2020radar}
A.~Kramer, C.~Stahoviak, A.~Santamaria-Navarro, A.-a. Agha-mohammadi, and C.~Heckman, ``Radar-inertial ego-velocity estimation for visually degraded environments,'' in \emph{2020 IEEE International Conference on Robotics and Automation (ICRA)}, 2020, pp. 5739--5746.

\bibitem{Huang2024Less}
Q.~Huang, Y.~Liang, Z.~Qiao, S.~Shen, and H.~Yin, ``Less is more: Physical-enhanced radar-inertial odometry,'' in \emph{2024 IEEE international conference on robotics and automation (ICRA)}.\hskip 1em plus 0.5em minus 0.4em\relax IEEE, 2024.

\bibitem{zhuang20234d}
Y.~Zhuang, B.~Wang, J.~Huai, and M.~Li, ``4d iriom: 4d imaging radar inertial odometry and mapping,'' \emph{IEEE Robotics and Automation Letters}, 2023.

\bibitem{kubelka2023we}
V.~Kubelka, E.~Fritz, and M.~Magnusson, ``Do we need scan-matching in radar odometry?'' \emph{arXiv preprint arXiv:2310.18117}, 2023.

\bibitem{kramer2021coloradar}
A.~Kramer, K.~Harlow, C.~Williams, and C.~Heckman, ``Coloradar: The direct 3d millimeter wave radar dataset,'' 2021.

\bibitem{rodriguez2018importance}
M.~L. Rodr{\'\i}guez-Ar{\'e}valo, J.~Neira, and J.~A. Castellanos, ``On the importance of uncertainty representation in active slam,'' \emph{IEEE Transactions on Robotics}, vol.~34, no.~3, pp. 829--834, 2018.

\bibitem{yin2022dynam}
H.~Yin, S.~Li, Y.~Tao, J.~Guo, and B.~Huang, ``Dynam-slam: An accurate, robust stereo visual-inertial slam method in dynamic environments,'' \emph{IEEE Transactions on Robotics}, vol.~39, no.~1, pp. 289--308, 2022.

\bibitem{thrun2002probabilistic}
S.~Thrun, ``Probabilistic robotics,'' \emph{Communications of the ACM}, vol.~45, no.~3, pp. 52--57, 2002.

\bibitem{barfoot2024state}
T.~D. Barfoot, \emph{State estimation for robotics}.\hskip 1em plus 0.5em minus 0.4em\relax Cambridge University Press, 2024.

\bibitem{shan2020orcvio}
M.~Shan, Q.~Feng, and N.~Atanasov, ``Orcvio: Object residual constrained visual-inertial odometry,'' in \emph{2020 IEEE/RSJ International Conference on Intelligent Robots and Systems (IROS)}.\hskip 1em plus 0.5em minus 0.4em\relax IEEE, 2020, pp. 5104--5111.

\bibitem{merrill2022symmetry}
N.~Merrill, Y.~Guo, X.~Zuo, X.~Huang, S.~Leutenegger, X.~Peng, L.~Ren, and G.~Huang, ``Symmetry and uncertainty-aware object slam for 6dof object pose estimation,'' in \emph{Proceedings of the IEEE/CVF Conference on Computer Vision and Pattern Recognition}, 2022, pp. 14\,901--14\,910.

\bibitem{yuan2021pixel}
C.~Yuan, X.~Liu, X.~Hong, and F.~Zhang, ``Pixel-level extrinsic self calibration of high resolution lidar and camera in targetless environments,'' \emph{IEEE Robotics and Automation Letters}, vol.~6, no.~4, pp. 7517--7524, 2021.

\bibitem{jiang2022lidar}
B.~Jiang and S.~Shen, ``A lidar-inertial odometry with principled uncertainty modeling,'' in \emph{2022 IEEE/RSJ International Conference on Intelligent Robots and Systems (IROS)}.\hskip 1em plus 0.5em minus 0.4em\relax IEEE, 2022, pp. 13\,292--13\,299.

\bibitem{clark1999simultaneous}
S.~Clark and G.~Dissanayake, ``Simultaneous localization and map building using millimeter wave radar to extract natural features,'' in \emph{Proceedings 1999 IEEE International Conference on Robotics and Automation (Cat. No. 99CH36288C)}, vol.~2.\hskip 1em plus 0.5em minus 0.4em\relax IEEE, 1999, pp. 1316--1321.

\bibitem{doer2020ekf}
C.~Doer and G.~F. Trommer, ``An ekf based approach to radar inertial odometry,'' in \emph{2020 IEEE International Conference on Multisensor Fusion and Integration for Intelligent Systems (MFI)}.\hskip 1em plus 0.5em minus 0.4em\relax IEEE, 2020, pp. 152--159.

\bibitem{michalczyk2023multi}
J.~Michalczyk, R.~Jung, C.~Brommer, and S.~Weiss, ``Multi-state tightly-coupled ekf-based radar-inertial odometry with persistent landmarks,'' in \emph{2023 IEEE International Conference on Robotics and Automation (ICRA)}.\hskip 1em plus 0.5em minus 0.4em\relax IEEE, 2023, pp. 4011--4017.

\bibitem{lu2020milliego}
C.~X. Lu, M.~R.~U. Saputra, P.~Zhao, Y.~Almalioglu, P.~P. De~Gusmao, C.~Chen, K.~Sun, N.~Trigoni, and A.~Markham, ``milliego: single-chip mmwave radar aided egomotion estimation via deep sensor fusion,'' in \emph{Proceedings of the 18th Conference on Embedded Networked Sensor Systems}, 2020, pp. 109--122.

\bibitem{kellner2014instantaneous}
D.~Kellner, M.~Barjenbruch, J.~Klappstein, J.~Dickmann, and K.~Dietmayer, ``Instantaneous ego-motion estimation using multiple doppler radars,'' in \emph{2014 IEEE International Conference on Robotics and Automation (ICRA)}.\hskip 1em plus 0.5em minus 0.4em\relax IEEE, 2014, pp. 1592--1597.

\bibitem{retan2022radar}
K.~Retan, F.~Loshaj, and M.~Heizmann, ``Radar odometry on se(3) with constant acceleration motion prior and polar measurement model,'' 2022.

\bibitem{he2021kalman}
D.~He, W.~Xu, and F.~Zhang, ``Kalman filters on differentiable manifolds,'' \emph{arXiv preprint arXiv:2102.03804}, 2021.

\bibitem{pomerleau2015review}
F.~Pomerleau, F.~Colas, R.~Siegwart \emph{et~al.}, ``A review of point cloud registration algorithms for mobile robotics,'' \emph{Foundations and Trends{\textregistered} in Robotics}, vol.~4, no.~1, pp. 1--104, 2015.

\bibitem{4DRadarSLAM}
J.~Zhang, H.~Zhuge, Z.~Wu, G.~Peng, M.~Wen, Y.~Liu, and D.~Wang, ``4dradarslam: A 4d imaging radar slam system for large-scale environments based on pose graph optimization,'' in \emph{2023 IEEE International Conference on Robotics and Automation (ICRA)}, 2023, pp. 8333--8340.

\bibitem{yuan2022efficient}
C.~Yuan, W.~Xu, X.~Liu, X.~Hong, and F.~Zhang, ``Efficient and probabilistic adaptive voxel mapping for accurate online lidar odometry,'' \emph{IEEE Robotics and Automation Letters}, vol.~7, no.~3, pp. 8518--8525, 2022.

\bibitem{grupp2017evo}
M.~Grupp, ``evo: Python package for the evaluation of odometry and slam.'' \url{https://github.com/MichaelGrupp/evo}, 2017.

\bibitem{xu2021fast}
W.~Xu and F.~Zhang, ``Fast-lio: A fast, robust lidar-inertial odometry package by tightly-coupled iterated kalman filter,'' \emph{IEEE Robotics and Automation Letters}, vol.~6, no.~2, pp. 3317--3324, 2021.

\bibitem{harlow2023new}
K.~Harlow, H.~Jang, T.~D. Barfoot, A.~Kim, and C.~Heckman, ``A new wave in robotics: Survey on recent mmwave radar applications in robotics,'' \emph{arXiv preprint arXiv:2305.01135}, 2023.

\end{thebibliography}


\end{document}